\documentclass[lettersize, 10pt, journal, twoside]{IEEEtran}
\usepackage[T1]{fontenc}

\usepackage{amsmath,amssymb,amsfonts}
\usepackage{dsfont}
\usepackage{graphicx}
\usepackage{xcolor}
\usepackage{array}
\usepackage{textcomp}
\usepackage{multirow}
\usepackage{booktabs}
\usepackage{colortbl}
\usepackage{float}
\usepackage{pifont}

\usepackage{caption}
\usepackage{subcaption}
\usepackage{multirow,tabularx}
\usepackage{hyperref}
\usepackage{orcidlink}
\usepackage{algorithmic}
\usepackage{capt-of}

\usepackage{tikz}
\usetikzlibrary{arrows.meta,positioning,decorations.pathreplacing,calc,fit,shapes.geometric}

\usepackage{balance}
\usepackage{enumitem}
\usepackage[capitalise,noabbrev]{cleveref}
\hypersetup{colorlinks=true,linkcolor=blue!60!black,citecolor=green!50!black,urlcolor=blue!70!black}

\ifdefined\MARKED
  \colorlet{revcolor}{blue!55!black}
  \newcommand{\rev}[1]{\textcolor{revcolor}{#1}}
  \newenvironment{revblock}{\par\color{revcolor}}{\par}
\else
  \newcommand{\rev}[1]{#1}
  \newenvironment{revblock}{\par}{\par}
\fi
\ifdefined\FINMARKED
  \newcommand{\fin}[1]{\textcolor{red!70!black}{#1}}
\else
  \newcommand{\fin}[1]{#1}
\fi
\graphicspath{{figures/}}
\AtBeginDocument{\raggedbottom}
\newcommand{\method}{\textsc{SafeVPR}}
\newcommand{\fdr}{\mathrm{FDR}}

\newcommand{\dinov}{DINOv2}

\title{
    \textsc{SafeVPR}: Patch-Based Verification with \rev{Empirical FDR Control} \\
    for Cross-Condition Sequence Visual Place Recognition
}

\author{\phantom{Anonymous Authors}}
\author{
    Ha Sier\,\orcidlink{0009-0000-3617-107X},\,
    Jiaqiang Zhang\,\orcidlink{0000-0002-4509-8115},\,
    Zhuo~Zou\,\orcidlink{0000-0002-8546-1329},\,
    Xianjia~Yu\,\orcidlink{0000-0002-9042-3730},\,
    Tomi~Westerlund\,\orcidlink{0000-0002-1793-2694}%
\thanks{This research is supported by the Research Council of Finland's Digital
Waters (DIWA) flagship (Grant No. 359247) and the DIWA Doctoral
Training Pilot project funded by the Ministry of Education and Culture
(Finland).}
\thanks{Corresponding authors: Ha Sier and Xianjia Yu.}%
\thanks{Ha Sier, Jiaqiang Zhang, Xianjia Yu, and Tomi Westerlund are with \href{https://tiers.utu.fi}{Turku Intelligent Embedded and Robotic Systems (TIERS) Lab}, University of Turku, Turku, Finland, e-mail: {\tt\footnotesize \{sierha, jiaqiang.zhang, xianjia.yu, tovewe\}@utu.fi.}}%
\thanks{Zhuo Zou is with the School of Information Science and Technology, Fudan University, Shanghai, China. E-mail: {\tt\footnotesize zhuo@fudan.edu.cn}.}%
}

\begin{document}

\maketitle

\begin{abstract}
Sequence-based visual place recognition (VPR) for loop closure and robot
relocalization must decide whether the retrieved top-1 candidate should be
accepted. Conformal calibration can set this accept/reject threshold, but its
finite-sample guarantees require exchangeability between calibration
and deployment, a requirement that cross-condition deployment violates. \rev{We therefore claim no formal guarantee
under condition shift; this paper instead asks which verification score and
calibration design keep a target false-discovery rate (FDR) \emph{empirically}
satisfied across realistic appearance change.} We present \method, a
non-trainable pipeline that replaces the backbone cosine similarity with a
mutual-nearest-neighbour (MNN) patch score from frozen \dinov{} features,
and flat Learn-Then-Test (LTT) calibration with \fin{Mondrian (per-bin)
Clopper--Pearson thresholds}. Across 23
cross-condition \rev{ground-vehicle} setups (Oxford RobotCar, NCLT, St~Lucia; three
frozen backbones), \method\ meets the FDR target on 23/23 setups at
$\alpha=0.10$ (mean accepted FDR 0.014, mean TPR 0.75) \rev{and on 22/23 at
$\alpha=0.05$ with 22 retaining non-trivial coverage (best cosine
variant\fin{, cosine$+$LTT}: 22/23 valid but 17 non-trivial); under leave-one-dataset-out
calibration, validity drops to 20/23 with 14 retaining useful coverage
(strongest cosine baseline: 16/23 and 10), and on texture-poor scenery (Nordland)
the verifier is uninformative and the system abstains entirely\fin{, as it
does everywhere at $\alpha=0.01$}; we report
these as limitations. In a pose-graph relocalization study, gating loop-closure
constraints reduces erroneous constraints by $29.6\times$ and cuts
the median trajectory error from 95.5\,m to 9.7\,m.}
Code is available at \url{https://github.com/TIERS/SafeVPR}.
\end{abstract}

\begin{IEEEkeywords}
Visual place recognition, conformal prediction, false discovery rate,
foundation models, calibration under distribution shift, \rev{risk-aware robot
localization}, condition-robust retrieval verification
\end{IEEEkeywords}

\IEEEpeerreviewmaketitle

\section{Introduction}\label{sec:intro}

\IEEEPARstart{S}equence-based visual place recognition (VPR) drives
loop closure, map reuse, and relocalization. Retrieval alone,
however, is not enough: once a top-1 sequence is retrieved, the
robot must decide whether to accept it as a localization constraint.
The decision is asymmetric: an accepted false match can corrupt the
map, while a rejected true match merely delays relocalization. It
therefore requires calibrated risk control, not only retrieval
accuracy.

Conformal prediction~\cite{vovk2005algorithmic} offers the natural
framework: Learn-Then-Test (LTT)~\cite{angelopoulos2021learn}
selects a threshold whose accepted predictions meet a user-specified
false-discovery-rate (FDR) target with finite-sample confidence,
\emph{provided} calibration and deployment are exchangeable.
Cross-condition VPR breaks exactly this: a robot calibrated in one
condition and deployed under different lighting, weather, or season
no longer has exchangeable scores, and no conformal procedure can
formally guarantee FDR under arbitrary unseen shift. The practical
question is how to design the score and the calibration procedure so
that \emph{empirical} FDR validity is recovered across realistic
condition changes. \rev{Our claim is correspondingly empirical, on a fixed
grid of held-out ground-vehicle deployments; transfer to unseen
datasets is measured, not assumed (\Cref{sec:results}).}

The direct baseline, one LTT threshold on the backbone cosine
similarity, fails to control FDR on $9/23$ cross-condition setups
at $\alpha=0.10$, for two interacting reasons: the cosine score is
condition-dependent (its distribution shifts between calibration and
deployment), and a single global threshold treats confident and
ambiguous matches identically although their reliability shifts
differently. \method\ replaces both pieces with non-trainable
components. The cosine score gives way to a bounded verification
score: the fraction of frozen \dinov{}~\cite{oquab2024dinov2}
patches that survive mutual-nearest-neighbour (MNN) matching and a
Lowe-ratio test between query and candidate. The global threshold
gives way to Mondrian conformal LTT~\cite{vovk2003mondrian}, which
fits a separate \fin{Clopper--Pearson-screened} threshold per score bin. Nothing
is trained; only the conformal thresholds are fitted.

Across $23$ cross-condition setups spanning three datasets (Oxford
RobotCar~\cite{maddern2017oxford}, NCLT~\cite{carlevaris2016nclt},
St~Lucia~\cite{glover2010stlucia,berton2022deepvg}) and three frozen
VPR backbones (NetVLAD~\cite{arandjelovic2016netvlad},
CosPlace~\cite{berton2022cosplace}, D$^2$-VPR~\cite{lu2024d2vpr}),
\method\ is empirically valid on all $23$ setups at $\alpha=0.10$
(mean accepted FDR $0.014$, mean TPR $0.75$). The central finding:
raw discrimination does not buy validity. AnyLoc-VLAD ties our
AUROC yet fails more setups under the same calibration,
SuperPoint$+$LightGlue attains the highest recall while losing
validity where its score shifts, and learned cross-attention
re-ranking fares worse still.

The contributions of this work are as follows:
\begin{enumerate}[leftmargin=1.3em,itemsep=1pt, label=\roman*).]
    \item We introduce \method, a non-trainable
    verification-and-calibration pipeline for cross-condition sequence
    VPR: frozen \dinov{} patch-MNN verification with Mondrian conformal
    LTT.
    \item We demonstrate empirical FDR validity on $23/23$
    cross-condition setups (three datasets $\times$ three backbones),
    persisting under calibration resampling, condition hold-out, and
    $5$\,m metric pose-error targets, \rev{and report the boundary
    transparently: graceful degradation (validity retained, coverage
    lost) under leave-one-dataset-out deployment, and global
    abstention on repetitive scenery where the verifier is
    uninformative \fin{or at targets as strict as $\alpha=0.01$}.}
    \item We show, across six alternative verifiers, that retrieval
    discrimination does not imply empirical validity, \rev{and trace the
    cause to the cross-condition stability of the score-conditional
    error map.}
    \item \rev{We evaluate the decision end-to-end in a pose-graph
    relocalization study on odometry from the datasets' real sensor
    streams: gating with \method\ reduces erroneous constraints by
    $29.6\times$ and median trajectory error by $9.8\times$, across
    all $23$ setups.}
\end{enumerate}

\section{Related Work}\label{sec:related}

\subsection{Foundation-model patches in VPR}

\dinov{}~\cite{oquab2024dinov2} patch features transfer robustly
across appearance change, and recent VPR systems exploit them for
retrieval and re-ranking: AnyLoc~\cite{keetha2024anyloc} aggregates
frozen patches with unsupervised VLAD/GeM pooling,
EffoVPR~\cite{tziony2024effovpr} re-ranks with self-attention
features, and SelaVPR~\cite{lu2024selavpr} adds lightweight adapters.
Before the foundation era,
Patch-NetVLAD~\cite{hausler2021patchnetvlad} introduced
mutual-nearest-neighbour (MNN) patch matching over regional NetVLAD
descriptors as a retrieval-quality signal. \method\ follows this
patch-level verification philosophy but differs in ingredients and,
more fundamentally, in selection criterion: frozen \dinov{} tokens in
place of task-specific descriptors, \rev{and a score selected not for
retrieval accuracy but for the cross-condition stability of its
\emph{conditional error map}, the property calibration transfer
actually requires (\Cref{sec:mechanism}). Patch-NetVLAD demonstrates
that MNN patch matching ranks well; we show \emph{which} patch
statistic calibrates stably, and why.} None of these works address the
acceptance-risk question: once a top-1 candidate is retrieved, should
it be accepted under an unseen condition? We use patch matching not
as another retrieval score but as the calibrated quantity inside a
conformal accept/reject procedure.

\subsection{Conformal prediction for visual retrieval}

Conformal prediction gives distribution-free guarantees under
exchangeability~\cite{vovk2005algorithmic};
Learn-Then-Test (LTT)~\cite{angelopoulos2021learn} extends it to
risk-controlled threshold selection (generalised
in~\cite{angelopoulos2024conformal}), and Mondrian
variants~\cite{vovk2003mondrian} calibrate per
category or score region. Applications to place recognition remain
limited~\cite{tellex2025partialcertainty}. \rev{Three adjacent frameworks
do not transfer to our setting.} Importance-weighted conformal
prediction~\cite{tibshirani2019conformal} handles covariate shift via
density ratios, but under severe condition shift the weights
concentrate and the calibrator collapses to near-total abstention
(\Cref{sec:learned_reranker}). \rev{Selective
prediction~\cite{geifman2017selective} and risk-controlling
prediction sets~\cite{bates2021distribution} assume deployment
matches the held-out distribution on which risk was estimated,
precisely what cross-condition VPR breaks. Adaptive conformal
methods~\cite{gibbs2021adaptive} track shift online but need
deployment-time label feedback, which a relocalizing robot lacks:
correctness of an accepted match is unknown at decision time. Our
setting therefore demands a score whose risk estimate transfers
\emph{without} labels or adaptation, the design question
\Cref{sec:mechanism} addresses.} \rev{We claim no formal guarantee under
arbitrary shift (exchangeability is violated); we study how far a
condition-robust verification score plus Mondrian LTT recovers
\emph{empirical} FDR validity.}

\subsection{Verification and learned re-rankers}

Retrieve-and-verify pipelines classically check candidates with
local features and geometric
consistency~\cite{lowe2004distinctive,sattler2017large}; learned
matchers such as LightGlue~\cite{lindenberger2023lightglue} are
natural VPR verifiers. But a strong verification score is not
automatically a calibration-stable one: in our experiments
LightGlue$+$RANSAC discriminates well on NCLT yet loses validity on
Oxford night and dusk, where keypoint detection and geometric
matching destabilise (\Cref{sec:verifier_sweep}). Learned
cross-attention re-rankers improve in-condition accuracy but their
output probabilities over-shoot under condition shift, and standard
mitigations do not recover FDR validity
(\Cref{sec:learned_reranker}). This motivates \method's
\emph{non-trainable} foundation-patch verifier: it avoids learning a
condition-dependent score while retaining patch-level matching
evidence.

\begin{figure*}[t!]
    \centering
    \includegraphics[width=\textwidth]{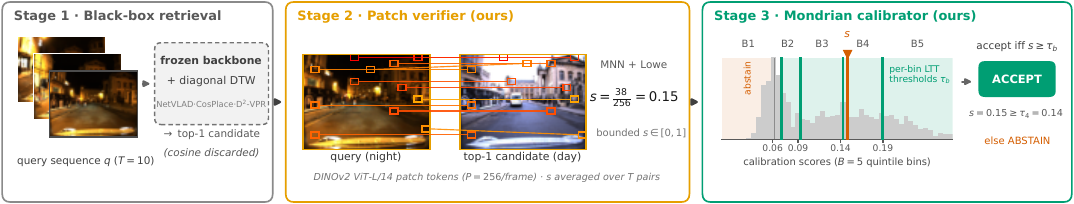}
    \caption{\textbf{The \method\ pipeline.} \emph{Stage 1} (grey,
    black box): a frozen VPR backbone and diagonal-DTW matcher return
    a top-$1$ candidate; the cosine score is discarded. \emph{Stage 2}
    (orange, ours): frozen \dinov{} patch MNN matching with a
    Lowe-ratio test gives a bounded score $s \in [0,1]$.
    \emph{Stage 3} (teal, ours): Mondrian calibration fits per-bin
    \fin{Clopper--Pearson-screened} thresholds $\tau_b$ over $B{=}5$
    score quintiles; accept iff $s \ge \tau_b$; infeasible bins abstain.}
    \label{fig:safevpr_pipeline}
\end{figure*}

\section{Methodology}\label{sec:method}

\subsection{Problem setup and conformal contract}\label{sec:problem_setup}

A query is a sequence of $T$ consecutive frames
$\mathbf{q} = (q_1, \dots, q_T)$. A frozen VPR
backbone~\cite{arandjelovic2016netvlad,berton2022cosplace,lu2024d2vpr}
maps each frame to a global descriptor; a frozen sequence matcher
scores every $T$-frame database window by diagonal-mean cosine,
re-ranks the top candidates with DTW, and returns the top-$1$
candidate $\mathbf{c}^\star$ with score $s_\mathrm{cos}\!\in\![0,1]$.
Both are black boxes; we observe only
$(\mathbf{c}^\star, s_\mathrm{cos})$ and the raw frames.

The system must \emph{accept} $\mathbf{c}^\star$ (returning the
candidate's pose) or \emph{reject} (returning ``unknown''). Given a
labelled calibration set $\{(\mathbf{q}_i, y_i)\}_{i=1}^n$ with
$y_i\!\in\!\{0,1\}$ marking top-$1$ correctness, we seek \fin{an
accept region $A$ (thresholds on a score $s$) whose
\emph{selective risk} $q(A) = \Pr(y=0 \mid s \in A)$ is at most
$\alpha$ with probability $\ge 1-\delta$ over the calibration draw.
For i.i.d.\ queries and fixed $A$ this bounds
$\fdr = \mathbb{E}[\mathrm{FDP}] = q(A)\Pr(\text{any accept})$, with
FDP the false fraction of accepts ($0$ if none). We report the
realized FDP, an estimate of the deployment selective risk, as
(empirical) FDR; empty accept sets are vacuously valid.} \method\ is a non-trainable verifier (the score
$s$, \Cref{sec:verifier}) plus a non-trainable calibrator
(\Cref{sec:calibrator}); \Cref{fig:safevpr_pipeline} summarises the
pipeline.

\subsection{Verifier: \dinov{} patch-MNN ratio}\label{sec:verifier}

Standard practice uses $s = s_\mathrm{cos}$, whose distribution moves
under condition shift. We replace it with a frozen foundation-model
verifier. For each diagonal frame pair $(q_t, c^\star_t)$ we extract
\dinov{} ViT-L/14~\cite{oquab2024dinov2} patch tokens
($P\!=\!256$ patches, $d\!=\!1024$, final layer, $224\!\times\!224$
input), $\ell_2$-normalise, form the patch cosine matrix
$\mathbf{S}^t$, and keep a query patch only if (i) its best match is
a \emph{mutual} nearest neighbour and (ii) it passes the Lowe ratio
test $S^t_{ij_2}/S^t_{ij_1} < r$ with $r=0.9$ (the runner-up must be
clearly weaker; ties reject). With $\rho_t$ the fraction of query
patches surviving both filters, the verification score is the
temporal mean $s = \frac{1}{T}\sum_{t=1}^T \rho_t$ ($T{=}10$):
bounded in $[0,1]$, training-free, and independent of VPR-specific
aggregation.

\subsection{Calibrator: Mondrian conformal LTT}\label{sec:calibrator}

We calibrate on $\{(s_i, y_i)\}_{i=1}^n$ with $\delta=0.05$ and
user-chosen $\alpha\!\in\!\{0.05, 0.10\}$. Vanilla
LTT~\cite{angelopoulos2021learn} tests $M=5$ candidate thresholds
(empirical quantiles of the calibration scores) and keeps the
\fin{lowest one (largest accept set)} whose Clopper--Pearson upper
bound on the accepted false rate, at confidence $1-\delta/M$, is at
most $\alpha$; a finer
grid buys little because only the clean high-score region can pass
the bound. Mondrian conformal~\cite{vovk2003mondrian} relaxes the
single-threshold assumption: we bin calibration queries into $B=5$
quintiles of $s$, \fin{and within each bin screen every threshold:
each prefix of the bin's score-sorted queries is bounded by
Clopper--Pearson at confidence $1-\delta/B$, and $\tau_b$ is the
score of the largest passing prefix.} A test query is routed to its
bin via the calibration-derived edges and accepted iff
$s_\star \geq \tau_{b(s_\star)}$; bins with no passing threshold
abstain ($\tau_b = +\infty$). \fin{This screen is a
heuristic: the threshold scan is uncorrected, bins and thresholds
are fitted on the tested data, and prefix counts omit ties that
$s \ge \tau_b$ accepts. A split-then-certify variant fixes bins and
ten candidates per bin on $30\%$ of the pool and certifies each
candidate's full accept set on the rest (Bonferroni over $10B$); it
carries the statement above under exchangeability and keeps the
headline ($23/23$ at $\alpha=0.10$, all non-trivial, TPR $0.67$
vs.\ $0.75$; full rerun in the code repository). Reported results use
the screen.} When $n_\text{cal} < 5B/\alpha$ the
recipe falls back to vanilla LTT; \rev{on the $23$-setup grid this never
triggers ($n_\text{cal} \in [1393, 1887]$ vs.\ thresholds
$250$/$500$), and \Cref{sec:limits} quantifies the calibration
budget at which it would.} All components except $\{\tau_b\}$ are
frozen; inference is two \dinov{} forwards per query, a
$P\!\times\!P$ cosine, and a two-comparison lookup.

\begin{revblock}
\subsection{Assumptions and scope}\label{sec:assumptions}

Under
exchangeability of calibration and deployment scores, \fin{the
certified variant yields the $(\alpha,\delta)$ statement of
\Cref{sec:problem_setup}; the screen does not formally.} Cross-condition
deployment violates exchangeability, and Mondrian binning does
\emph{not} restore it: quantile bins are fitted on calibration
scores, and nothing prevents the deployment distribution from
occupying them differently. Consequently, none of the results in
this paper constitute a distribution-free guarantee.

What the procedure actually relies on is weaker and measurable. Let
$A$ be the accept region fitted at calibration, $q_{\mathrm{cal}} =
\Pr_{\mathrm{cal}}(\text{false} \mid A)$ \fin{its population
selective risk on the calibration distribution, which the
calibrator's bound targets ($q_{\mathrm{cal}} \le \alpha$ holds with
confidence $1-\delta$, not by construction)}, and $q_{\mathrm{dep}}$
the same quantity under deployment. \fin{Trivially,}
\begin{equation}
\fin{q_{\mathrm{dep}}} \;=\; q_{\mathrm{cal}} + \varepsilon,
\qquad \varepsilon \;=\; q_{\mathrm{dep}} - q_{\mathrm{cal}},
\label{eq:eps}
\end{equation}
so the target is exceeded \fin{when} the \emph{conditional}
false-match rate over the accept region drifts by more than the
\fin{risk slack $\alpha - q_{\mathrm{cal}}$ (not the confidence
budget $\delta/B$). We estimate $\varepsilon$ as the difference of
realized false fractions on $A$ (deployment minus calibration).} The operative assumption is therefore not
exchangeability of the score distribution but stability of the
score-conditional error map on the accept region --- a property of
the verification score, measurable on held-out deployments.
Empirically (\Cref{sec:mechanism}), the worst case of $\varepsilon$
across all $23$ setups and both $\alpha$ levels is $+0.049$ for the
patch-MNN score, against $+0.358$ for cosine; this difference, not
any global distribution-distance argument, \fin{tracks} the validity
results. \Cref{eq:eps} \fin{decomposes population
risks; it is not a finite-sample identity for realized errors and}
certifies nothing in advance, but it isolates the single quantity
a deployment must keep small; \Cref{sec:robustness} probes its
limits (leave-one-dataset-out) and \Cref{sec:discussion} reports
where it fails (Nordland).

Scope: all claims are empirical and ground-vehicle; transfer to
unseen datasets is measured by the LODO protocol
(\Cref{sec:robustness}), and platform boundaries are drawn in
\Cref{sec:discussion}.
\end{revblock}

\section{Experimental Setup}\label{sec:setup}

\subsection{Datasets, backbones, protocol}
Oxford RobotCar~\cite{maddern2017oxford},
NCLT~\cite{carlevaris2016nclt}, and St~Lucia~\cite{glover2010stlucia,
berton2022deepvg} span urban night/weather, campus foliage, and
time-of-day shifts; \rev{Nordland is excluded from the grid and analysed
separately as a failure mode (\Cref{sec:discussion})}.
Three frozen VPR backbones: NetVLAD (ResNet-18, principal component
analysis [PCA] to 256-D), CosPlace
(ResNet-18, $D\!=\!512$), and D$^2$-VPR (\dinov{}-small with VPR head,
PCA-256), spanning a strength range from R@1 $\sim\!0.2$
(NetVLAD-Oxford-night) to $\sim\!0.97$ (D$^2$-VPR). NetVLAD lacks
NCLT-2013-04-05, giving $8\!\times\!3 - 1 = 23$ setups. For each
held-out condition $h$ we calibrate on the union of the other
conditions' calibration slices (per backbone) and test on the entire
held-out condition. Calibration / test queries are split using a
buffer-based protocol: each dataset's database is segmented along
arc-length into a [train\,$|$\,buffer\,$|$\,eval\,$|$\,buffer] cycle,
with a post-filter that drops eval queries within $50\,\mathrm{m}$ of
any train query in 2D space. This eliminates the route-loop leakage
that naive frame-index splits suffer. \fin{Frames are
distance-subsampled ($2$\,m; $\approx\!12$\,m on St~Lucia) and a
$T{=}10$ window starts at every frame, so neighbours share nine
frames. Windows are assigned by centre, and the $100$--$200$\,m
buffers exceed a window's half-span, so calibration windows never
touch the (for \method\ unused) train segment; calibration and test
never share a traversal. The $23$ setups share $8$ query traversals,
so they are not independent. Frames are resized to $224{\times}224$
before \dinov{} (native $640{\times}480$ to $1616{\times}1232$),
fixing the patch grid; only LODO mixes cameras.}

\subsection{Metrics}
A setup is \emph{valid} at level $\alpha$ if its \fin{realized FDP
(reported as FDR, \Cref{sec:problem_setup})} is at most $\alpha$ (empty accept sets count as vacuously valid),
and \emph{non-trivially valid} if it accepts $>5\%$ of test queries.
We report mean accepted-FDR and TPR, aggregated over setups with
non-empty accept sets. To probe calibration stability we additionally
report \emph{robust-pass}: a setup is robust-pass if at least $95\%$ of
its calibration resamples remain valid. AUROC quantifies a score's
raw discrimination independent of any threshold, and a \emph{metric}
validity variant replaces the correctness label with
$|\text{pose error}|\!\leq\!\tau_p$.

\section{Results}\label{sec:results}

\subsection{Headline cross-condition result}\label{sec:headline}

\Cref{tab:cross_main} reports \method\ against cosine\,$+$\,LTT and the
two intermediate variants (\Cref{sec:setup}). \method\ achieves
$\mathbf{23/23}$ valid setups
at $\alpha=0.10$ with every setup providing non-trivial coverage,
mean accepted FDR $0.014$ \fin{($95\%$ CI $[0.008, 0.023]$; paired temporal block
bootstrap, $100$-window $\approx\!200$\,m blocks shared by a
traversal's three backbones, conditional on the fixed grid and
calibration pools)}, and mean TPR $0.75$ \fin{($[0.717, 0.780]$).
Cosine$+$Mondrian stays disjoint (FDR $[0.026, 0.049]$, TPR
$[0.582, 0.635]$); the paired differences exclude zero for blocks
of $1$ to $100$ windows.} \rev{At $\alpha=0.10$ the two
design axes are not yet separated: LTT $\to$ Mondrian under cosine
lifts $14\!\to\!22$ valid (the NCLT recovery is the calibrator's)
and cosine $\to$ MNN under flat LTT lifts $14\!\to\!20$. Every
stricter probe separates them in the verifier's favour: $22/23$
vs.\ $19/23$ at $\alpha=0.05$; $20/23$ valid with $14$ non-trivial
vs.\ $16/23$ with $10$ under leave-one-dataset-out calibration
(\Cref{sec:robustness}); worst accept-region drift $+0.049$ vs.\
$+0.358$ (\Cref{sec:mechanism}); $\sim\!14$\,pp more TPR at matched
validity; and fewer erroneous pose-graph constraints at higher
coverage (\Cref{sec:slam_gate}). Without a GPU, cosine$+$Mondrian
remains a fallback at the loose operating point.}

\begin{table*}[t]
\centering
\caption{Cross-condition FDR validity across the $23$ setups; the held-out condition is never seen in calibration. ``Pass'': empirical FDR $\leq \alpha$; ``Non-tr.'': additionally coverage $> 5\%$. Means: FDR over non-empty accept sets, TPR over all setups. \textbf{Bold}: per-block best FDR/TPR. Red tint deepens with failure severity; a \colorbox{red!15}{tinted TPR cell} marks best recall despite failed validity.}
\label{tab:cross_main}
\footnotesize
\setlength{\tabcolsep}{4pt}
\begin{tabular*}{\linewidth}{@{\extracolsep{\fill}}l rrrr rrrr}
\toprule
& \multicolumn{4}{c}{$\alpha = 0.05$} & \multicolumn{4}{c}{$\alpha = 0.10$} \\
\cmidrule(lr){2-5}\cmidrule(lr){6-9}
Method & Pass & Non-tr. & FDR\,$\downarrow$ & TPR\,$\uparrow$ & Pass & Non-tr. & FDR\,$\downarrow$ & TPR\,$\uparrow$ \\
\midrule
\multicolumn{9}{l}{\emph{Oxford ($4$ traversals $\times 3$ backbones, $12$ setups)}} \\
\quad cosine $+$ LTT                       & 12/12                       & \cellcolor{red!18}8/12      & \textbf{0.0000} & 0.380                            & 12/12                       & \cellcolor{red!18}9/12      & 0.0005          & 0.453                            \\
\quad cosine $+$ Mondrian                  & 12/12                       & \cellcolor{red!18}7/12      & \textbf{0.0000} & 0.390                            & 12/12                       & \cellcolor{red!18}9/12      & \textbf{0.0000} & 0.516                            \\
\quad \dinov{} MNN $+$ LTT                 & \cellcolor{red!7}11/12      & \cellcolor{red!7}10/12      & 0.0097          & 0.839                            & \cellcolor{red!7}11/12      & \cellcolor{red!7}11/12      & 0.0508          & \textbf{0.944}                   \\
\quad \textbf{\textsc{SafeVPR} (ours)}     & 12/12                       & 12/12                       & 0.0036          & \textbf{0.857}                   & 12/12                       & 12/12                       & 0.0093          & 0.916                            \\
\midrule
\multicolumn{9}{l}{\emph{NCLT ($3$ sessions $\times 3$ backbones $-1$ missing, $8$ setups)}} \\
\quad cosine $+$ LTT                       & 8/8                         & \cellcolor{red!7}7/8        & 0.0105          & 0.423                            & \cellcolor{red!32}0/8       & \cellcolor{red!32}0/8       & 0.1921          & \cellcolor{red!15}\textbf{0.800} \\
\quad cosine $+$ Mondrian                  & \cellcolor{red!18}5/8       & \cellcolor{red!18}4/8       & 0.0331          & \cellcolor{red!15}\textbf{0.562} & 8/8                         & 8/8                         & 0.0486          & 0.646                            \\
\quad \dinov{} MNN $+$ LTT                 & 8/8                         & \cellcolor{red!18}6/8       & \textbf{0.0079} & 0.318                            & \cellcolor{red!18}6/8       & \cellcolor{red!18}4/8       & 0.0944          & 0.480                            \\
\quad \textbf{\textsc{SafeVPR} (ours)}     & \cellcolor{red!7}7/8        & \cellcolor{red!7}7/8        & 0.0131          & 0.402                            & 8/8                         & 8/8                         & \textbf{0.0232} & 0.490                            \\
\midrule
\multicolumn{9}{l}{\emph{St~Lucia ($1$ day-pair $\times 3$ backbones, $3$ setups)}} \\
\quad cosine $+$ LTT                       & \cellcolor{red!18}2/3       & \cellcolor{red!18}2/3       & 0.0873          & 0.796                            & \cellcolor{red!18}2/3       & \cellcolor{red!18}2/3       & 0.1539          & \cellcolor{red!15}\textbf{0.964} \\
\quad cosine $+$ Mondrian                  & \cellcolor{red!18}2/3       & \cellcolor{red!18}2/3       & 0.1193          & \cellcolor{red!15}\textbf{0.872} & \cellcolor{red!18}2/3       & \cellcolor{red!18}2/3       & 0.1273          & 0.890                            \\
\quad \dinov{} MNN $+$ LTT                 & 3/3                         & 3/3                         & \textbf{0.0015} & 0.677                            & 3/3                         & 3/3                         & \textbf{0.0015} & 0.677                            \\
\quad \textbf{\textsc{SafeVPR} (ours)}     & 3/3                         & 3/3                         & 0.0116          & 0.658                            & 3/3                         & 3/3                         & 0.0115          & 0.780                            \\
\midrule
\multicolumn{9}{l}{\emph{All ($23$ setups): headline}} \\
\quad cosine $+$ LTT                       & \cellcolor{red!7}22/23      & \cellcolor{red!18}17/23     & 0.0182          & 0.449                            & \cellcolor{red!18}14/23     & \cellcolor{red!18}11/23     & 0.0954          & 0.641                            \\
\quad cosine $+$ Mondrian                  & \cellcolor{red!7}19/23      & \cellcolor{red!18}13/23     & 0.0311          & 0.513                            & \cellcolor{red!7}22/23      & \cellcolor{red!7}19/23      & 0.0350          & 0.610                            \\
\quad \dinov{} MNN $+$ LTT                 & \cellcolor{red!7}22/23      & \cellcolor{red!7}19/23      & \textbf{0.0080} & 0.637                            & \cellcolor{red!7}20/23      & \cellcolor{red!7}18/23      & 0.0595          & 0.748                            \\
\quad \textbf{\textsc{SafeVPR} (ours)}     & \cellcolor{red!7}22/23      & \cellcolor{red!7}22/23      & \textbf{0.0080} & \textbf{0.673}                   & 23/23                       & 23/23                       & \textbf{0.0144} & \textbf{0.750}                   \\
\bottomrule
\end{tabular*}
\end{table*}

The dataset sub-blocks in \Cref{tab:cross_main} show the headline
is not evenly hard. On Oxford, cosine$+$LTT already controls FDR but
with weak coverage (mean TPR $\sim\!0.4$); \method\ recovers
$12/12$ non-trivial and roughly doubles TPR. NCLT is where the
calibrator carries the result (cosine$+$LTT collapses to $0/8$ at
$\alpha=0.10$; both Mondrian variants recover). St~Lucia's mild
shift needs \method\ only for \texttt{st\_lucia\_netvlad}, where
the weak backbone leaves no cosine margin.

\rev{\emph{The calibrator buys validity at the loose operating point; the verifier buys everything that survives tightening.}}

\subsection{Verifier sweep: discrimination decouples from validity}\label{sec:verifier_sweep}

\Cref{tab:verifier_sweep} compares \method's score against three
within-patch aggregations of the same cached \dinov{} patches
(\texttt{patch\_mean}, \texttt{patch\_max}, \texttt{patch\_top10}) and
two published external verifiers: AnyLoc-VLAD~\cite{keetha2024anyloc}
and SuperPoint$+$LightGlue$+$RANSAC~\cite{lindenberger2023lightglue}
(mean diagonal-pair cosine and geometric inlier ratio,
respectively).

Per-setup mean AUROC is $0.948$ for MNN$+$Lowe and a statistically
indistinguishable $0.947$ for AnyLoc (patch aggregations
$0.81$--$0.88$; LightGlue $0.851$ but strongly dataset-dependent).
Yet validity decouples sharply: only \method\ achieves $20/20$ at
$\alpha=0.10$ with non-trivial coverage, AnyLoc reaches $15/20$
under Mondrian (TPR $0.47$), and LightGlue$+$LTT posts the highest
mean TPR of any method ($0.79$) at only $16/20$ valid, trading
validity for recall on Oxford.

\begin{table}[t]
\centering
\caption{Verifier $\times$ calibrator sweep ($20$-setup
Oxford$+$NCLT grid, $\alpha=0.10$; no St~Lucia features for
AnyLoc/LightGlue). Columns as in \Cref{tab:cross_main};
\rev{KS $=$ mean per-cell cal/test Kolmogorov--Smirnov distance}.
Italics: accept-every-top-$1$ baseline.}
\label{tab:verifier_sweep}
\scriptsize
\setlength{\tabcolsep}{3pt}
\begin{tabular*}{\linewidth}{@{\extracolsep{\fill}}ll rrrr}
\toprule
Score & Calibrator & Pass & Non-tr. & FDR & TPR \\
\midrule
  \textbf{MNN+Lowe} (ours)\rev{, KS $0.53$} & LTT & 17/20 & 15/20 & 0.0682 & 0.758 \\
   & \textbf{Mondrian} & \textbf{20/20} & \textbf{20/20} & \textbf{0.0148} & \textbf{0.745} \\
  \midrule
  AnyLoc-VLAD~\cite{keetha2024anyloc}\rev{, KS $0.50$} & LTT & 17/20 & 15/20 & 0.0612 & 0.709 \\
   & Mondrian & 15/20 & 11/20 & 0.0546 & 0.468 \\
  \midrule
  LightGlue+RANSAC~\cite{lindenberger2023lightglue}\rev{, KS $0.31$} & LTT & 16/20 & 16/20 & 0.0979 & 0.792 \\
   & Mondrian & 17/20 & 17/20 & 0.0477 & 0.716 \\
  \midrule
  DINOv2 patch mean & LTT & 11/20 & 3/20 & 0.1780 & 0.339 \\
   & Mondrian & 9/20 & 4/20 & 0.1466 & 0.283 \\
  \midrule
  Cosine top-1\rev{, KS $0.41$} & LTT & 12/20 & 9/20 & 0.0857 & 0.592 \\
   & Mondrian & 20/20 & 17/20 & 0.0205 & 0.568 \\
  \midrule
  \emph{Pure VPR (no calib.)} & \emph{---} & \emph{9/20} & \emph{9/20} & \emph{0.2363} & \emph{1.000} \\
\bottomrule
\end{tabular*}
\end{table}

Distribution distance does not predict validity either \rev{(KS
annotations in \Cref{tab:verifier_sweep})}: by mean per-cell cal/test
Kolmogorov--Smirnov distance our score does \emph{not} shift least
(LightGlue does, $0.31$ vs.\ our $0.53$), yet \method\
achieves $20/20$ where LightGlue reaches $17/20$: despite the
smallest overall shift, LightGlue leaks Oxford-night false matches
into its top \fin{accepting} bin. Only the accepted region's error
rate matters for validity; \Cref{sec:mechanism} \fin{examines} this
mechanism.

\rev{\emph{Choose the verification score by the stability of its accepted-region error rate, not by AUROC.}}

\begin{figure*}[t]
\centering
\begin{subfigure}[t]{0.693\textwidth}
\vspace{0pt}\centering
\includegraphics[width=\linewidth]{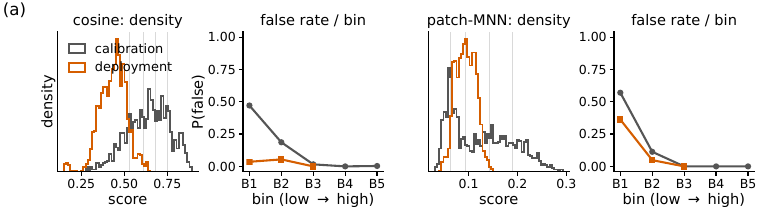}%
\refstepcounter{subfigure}\label{fig:mech_bins}%
\end{subfigure}\hfill
\begin{subfigure}[t]{0.283\textwidth}
\vspace{0pt}\centering
\includegraphics[width=\linewidth]{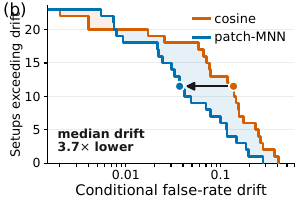}%
\refstepcounter{subfigure}\label{fig:mech_drift}%
\end{subfigure}
\caption{Mechanism of calibration transfer.
(a)~Oxford-night (D$^2$-VPR): score densities (cal vs.\ deployment;
grey verticals: bin edges) and each score's per-bin false-match
rate: cosine's fitted error map breaks at deployment; the bounded
patch-MNN's stays aligned.
(b)~Per-setup occupancy-weighted false-rate drift as exceedance
curves (left is stabler); shading: patch-MNN's margin.}
\label{fig:mechanism}
\end{figure*}

\begin{revblock}
\subsection{Mechanism: conditional stability of the verification score}
\label{sec:mechanism}

Why does the patch-MNN score calibrate stably where stronger-AUROC
scores do not? Mondrian LTT transfers a per-bin error model, so the
operative quantity is the \emph{conditional} false-match rate
$\Pr(\text{false} \mid \text{bin})$: calibration transfers
when this map is invariant \fin{on the accepted part of each bin},
however the marginal distribution moves
(\Cref{sec:assumptions}).

\Cref{fig:mech_bins} dissects the hardest Oxford condition: the
cosine density shifts wholesale (deployment mass concentrates in the
two lowest bins, whose fitted error rates no longer describe it,
and none reaches the accepting bins), while the bounded patch-MNN
ratio keeps density and per-bin false-rate map aligned, the
property Mondrian actually needs. \Cref{fig:mech_drift} quantifies this across the
grid with the deployment-occupancy-weighted drift of the per-bin
false rate: median $0.037$ for patch-MNN versus $0.137$ for cosine
($3.7\times$ lower; smaller on $17/23$ setups; drift above $0.15$
on $3$ setups versus cosine's $9$). The exceptions are informative:
St~Lucia's mild same-day shift leaves even the cosine map stable,
and the NCLT$\times$NetVLAD cells pair the weakest backbone with
the strongest seasonal aliasing. This foreshadows the
calibrator/verifier decomposition of \Cref{sec:robustness}.
The $\varepsilon$ of \Cref{eq:eps} is this drift restricted to the
accept region (the one that matters for validity): worst-case
$+0.049$ for patch-MNN versus $+0.358$ for cosine
(\Cref{sec:assumptions}), \fin{consistent with} the patch score
surviving the stricter $\alpha=0.05$ budget ($22/23$ vs.\ $19/23$).
\fin{Per-bin stability alone is not sufficient, as scores can move
within a bin; on the accepted subregions $\{b, s \ge \tau_b\}$
directly, the accept-weighted mean absolute drift is $0.010$ versus
$0.029$, exceeding $\alpha$ in $1/49$ subregions ($2\%$ of that
setup's accepts) versus $5/47$.}

\Cref{fig:mech_gallery} grounds the statistics in images. On the
aliased pair (third column), cosine assigns $0.42$, \emph{higher}
than the $0.36$ it assigns to a genuinely matching pair under the
same night-to-day change, while patch-MNN orders the two
correctly ($0.05$ vs.\ $0.15$): global descriptors average away the
evidence that distinguishes two similar streets, whereas
mutual-nearest-neighbour patch agreement with a ratio test requires
localized, spatially consistent correspondences. We also show the
failure cases the numbers imply: a rare accepted false match and a
conservative miss.

\begin{figure*}[t]

\centering
\begin{minipage}[t]{0.58\textwidth}
    \vspace{0pt}
    \centering
    \includegraphics[width=0.98\textwidth]{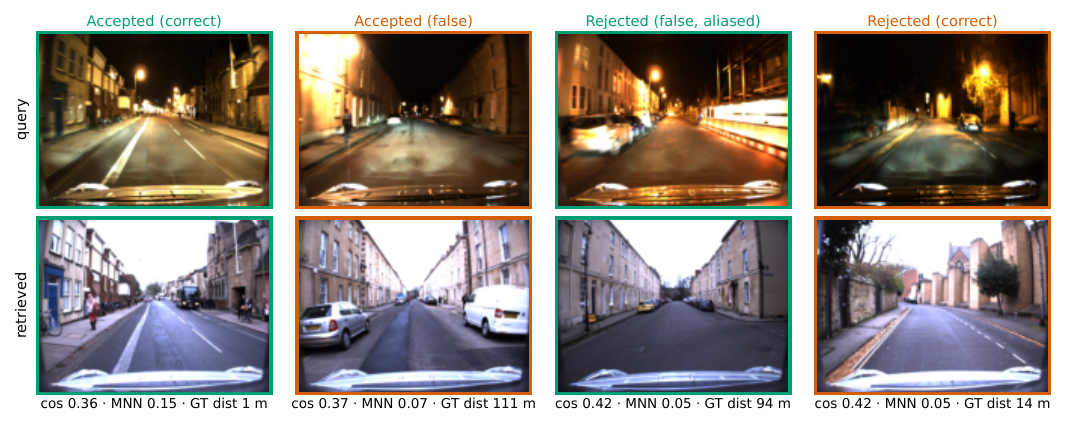}
    \caption{Decision cases on Oxford-night (CosPlace),
    $\alpha=0.10$\fin{; pose errors $0.7$, $111$, $94$, $14$\,m, left
    to right}. The cosine inversion: the aliased false pair (third
    column, cos $0.42$) outscores the correct pair (first, $0.36$);
    patch-MNN orders them correctly ($0.05$ vs.\ $0.15$). Second: a
    rare accepted false match; fourth: a conservative miss.}
    \label{fig:mech_gallery}
\end{minipage}
\hfill
\begin{minipage}[t]{0.40\textwidth}
\vspace{0pt}
\centering
\captionof{table}{Three robustness probes of increasing strictness
($23$-setup grid, $\alpha=0.10$ unless noted).
\emph{(a)~Bootstrap}: $500$ calibration resamples; robust-pass $=$
$\geq95\%$ valid resamples.
\emph{(b)~Hold-out}: drop one cal-pool condition at a time and
re-fit ($154$ valid pairs).
\emph{(c)~LODO}: calibrate on all datasets except the deployed one.
}
\label{tab:robustness}
\scriptsize
\setlength{\tabcolsep}{3pt}
\resizebox{\linewidth}{!}{
\begin{tabular}{l rrr}
\toprule
Probe (\method, $\alpha=0.10$) & Pass / N & Mean P(valid) / FDR & TPR \\
\midrule
\multicolumn{4}{l}{\emph{(a) Cal-sampling bootstrap}, per-setup} \\
  \quad robust-pass ($\geq 95\%$ resamples) & 22/23 & $0.996$ / $0.016$ & 0.734 \\
  \quad robust-pass at $\alpha = 0.05$       & 22/23 & $0.961$ / $0.008$ & 0.652 \\
\midrule
\multicolumn{4}{l}{\emph{(b) Cal-condition hold-out}, per pair} \\
  \quad pair pass                            & 151/154 & rate $0.981$ / $0.016$ & 0.734 \\
  \quad fully robust test setups             & 20/23 & --- / ---             & --- \\
  \quad pair pass at $\alpha = 0.05$         & 148/154 & rate $0.961$ / $0.006$ & 0.652 \\
\midrule
\multicolumn{4}{l}{\emph{(c) Leave-one-dataset-out (LODO)}} \\
  \quad empirical validity                   & 20/23 & --- / $0.046$ & 0.613 \\
  \quad non-trivial coverage                 & 14/23 & --- / ---     & --- \\
  \quad LODO at $\alpha = 0.05$              & 20/23 & --- / $0.045$ & 0.551 \\
\bottomrule
\end{tabular}
}

\centering

\end{minipage}

\end{figure*}

\emph{Patch-MNN calibrates stably \fin{where} its score-conditional error map transfers across conditions ($\varepsilon \le 0.049$), not because its marginal distribution shifts less.}
\end{revblock}

\subsection{Robustness: bootstrap, cal-condition hold-out, LODO}\label{sec:robustness}

\Cref{tab:robustness} reports the three calibration-perturbation probes
(\Cref{sec:setup}) in one table.

\emph{Cal-sampling bootstrap.}
Mean $P(\text{valid})\!=\!0.996$ at $\alpha=0.10$ ($0.961$ at
$0.05$), $22/23$ robust-pass at both levels; only
\texttt{nclt\_2013-04-05\_d2vpr} carries non-negligible mass above
the $\alpha=0.05$ line, matching the one empirical failure in
\Cref{tab:cross_main}.

\emph{Cal-condition hold-out.} $151/154$ valid $(c, c')$ pairs
($98.1\%$) remain valid at $\alpha=0.10$; $20/23$ test setups are
robust to \emph{every} single-condition removal. The three fragile
cells are the same that sat near the conformal boundary in the
bootstrap.

\emph{Cross-dataset (LODO).}
\rev{This is the strongest transfer test we run, the method's
honest boundary: with calibration drawn exclusively from
\emph{other} datasets, \method\ keeps
$\mathbf{20/23}$ setups valid at $\alpha=0.10$ but only $14/23$
retain non-trivial coverage (mean FDR $0.046$, mean TPR $0.61$);
on $9$ setups the system is valid mostly by abstention. The
verifier earns its keep here: cosine$+$Mondrian degrades to $16/23$ valid with
$10$ non-trivial ($\alpha=0.05$: $15/23$ and $6$, vs.\ \method's
unchanged $20/23$ and $14$). The three \method\ failures
(\texttt{nclt\_*\_d2vpr}) share one mechanism, a cross-dataset
$\varepsilon$-drift (\Cref{eq:eps}) of the \emph{lowest} bin:
calibrated on Oxford$+$St~Lucia only, urban B1 looks safe enough to
open ($\tau_1 = 0.047$, versus $\infty$ in-grid), and NCLT's
campus-foliage B1 ($92$--$94\%$ of deployment mass,
$\Pr(\text{false} \mid \mathrm{B1}) = 0.31$--$0.40$) floods in (FDR
$0.24$--$0.33$). The failure is directional (St~Lucia$+$NCLT $\to$
Oxford passes); a small target-domain calibration slice or
distribution-aware bin edges would close it, which we leave to
future work.}

\rev{\emph{Validity survives calibration perturbation; genuine cross-dataset transfer costs coverage and can open an unsafe low bin, so budget a small target-domain slice when changing datasets.}}

\subsection{Pose-error admissibility}

Replacing the correctness label with
$|\text{pose error}|\!\leq\!\tau_p$ and re-running the pipeline on
the densely-GPSed Oxford$+$NCLT subset ($20$ setups\fin{: all $12$
Oxford and $8$ NCLT setups, with calibration labels at the same
$\tau_p$}; St~Lucia's
$\sim\!5$\,m GPS subsampling makes sub-$25$\,m labels mechanically
unstable) leaves empirical metric-FDR control intact: $20/20$ valid
at $\alpha=0.10$ for every $\tau_p \in \{5,10,25\}$\,m. \rev{TPR is computed relative to the queries whose true pose error
is within $\tau_p$, so the $0.37$--$0.39$ TPR at $\tau_p{=}5$\,m
means the gate recovers $\approx\!38\%$ of genuinely sub-$5$\,m
matches: conservatism, not blindness.}

\rev{\emph{The recipe empirically controls metric error wherever the ground truth resolves it, and its TPR is measured against genuinely-close matches.}}

\begin{figure*}[t]
\centering
\includegraphics[scale=1]{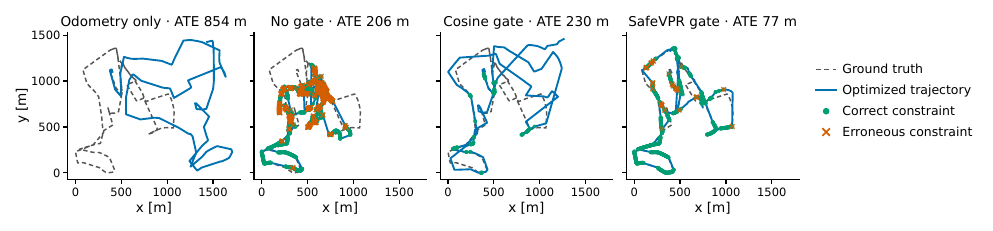}
\caption{Closed-loop relocalization on Oxford-night (CosPlace),
released stereo VO, $\alpha=0.10$. Ungated matching is dragged
toward aliased places; the cosine gate accepts too little to
contain drift here; \method\ keeps dense correct constraints and
tracks the ground truth.}
\label{fig:slam_traj}
\end{figure*}

\begin{revblock}
\subsection{Closed-loop relocalization study}\label{sec:slam_gate}

The FDR statements above concern the accept set in isolation. To test
whether the gated decision matters at the system level, we run a
pose-graph relocalization study on all $23$ setups: each query
traversal becomes a chain of SE(2) nodes connected by odometry
factors, and every \emph{accepted} top-1 match adds a $5$\,m position
prior at the matched database pose. On Oxford, odometry comes from the dataset's
released stereo VO~\cite{maddern2017oxford}; on NCLT, from the released wheel-encoder$+$FOG dead
reckoning~\cite{carlevaris2016nclt} ($100$\,Hz, $0.5\%$ drift over
$4.7$\,km); St~Lucia ships no odometry stream, so we compute
profile-based visual odometry from its released video, the
odometry model of the dataset's own release
paper~\cite{glover2010stlucia} ($7.8\%$ drift over $16.7$\,km),
calibrated on the database session's GPS only, so query ground truth
never enters the odometry chain. All $23$ setups thus run on
odometry from real sensor streams. Graphs are optimized with
Levenberg--Marquardt~\cite{gtsam} under five constraint policies:
odometry only, no gate (all top-1 matches), cosine$+$Mondrian gate,
\method, and a GT oracle.

\begin{table}[h]
\centering
\caption{Pose-graph optimization with gated loop-closure
constraints ($23$ setups, $\alpha=0.10$, real sensor-stream
odometry; \Cref{sec:slam_gate}). Accepted matches enter as
$5$\,m 2D position priors; ATE over per-setup RMSE.
FD $=$ erroneous constraints; Avail.\ $=$ accepted-query fraction;
Huber $=$ robust kernel.}
\label{tab:slam_gate}
\scriptsize
\setlength{\tabcolsep}{3.5pt}
\begin{tabular}{l rr rrr}
\toprule
& & & \multicolumn{3}{c}{ATE (m)} \\
\cmidrule(lr){4-6}
Constraint policy & FD & Avail. & median & mean & \begin{tabular}[b]{@{}r@{}}median\\(Huber)\end{tabular} \\
\midrule
Odometry only        & 0    & —    & 741.8 & 581.5 & 741.8 \\
No gate              & 5952 & 1.00 & 95.5  & 113.0 & 65.3  \\
Cosine$+$Mondrian    & 292  & 0.53 & 31.1  & 161.1 & \textbf{4.9}   \\
\textbf{\method}     & \textbf{201} & \textbf{0.63} & \textbf{9.7} & \textbf{49.9} & \textbf{4.9} \\
Oracle (upper bound) & 0    & 0.77 & 2.3   & 10.1  & 2.3   \\
\bottomrule
\end{tabular}
\end{table}

\Cref{tab:slam_gate} aggregates; \Cref{fig:slam_traj} shows the
hardest Oxford condition. Ungated matching injects $5952$ erroneous
constraints; \method\ admits $201$ ($29.6\times$ fewer) and cuts
median ATE from $95.5$\,m to $9.7$\,m. The cosine gate matches
\method\ in neither suppression ($292$ FD) nor coverage: where
odometry drifts, its sparse acceptances leave
long unconstrained stretches (on Oxford-night no better than no
gate, \Cref{fig:slam_traj}), inflating its mean ATE to $161$\,m.
\fin{On NCLT, where odometry is strong, each gate gives the lower ATE
on $4$ of the $8$ setups, with or without a robust kernel (per-setup
values in the code repository); across all $23$ the cosine gate
trails on every aggregate except the Huber median, where both reach
$4.9$\,m.} Strong odometry also
exposes a real boundary of gating alone: on $3/23$ setups the
wheel$+$FOG baseline is accurate enough ($18$--$63$\,m) that
\method's few residual false accepts leave the raw gated graph
\emph{above} odometry-only. A Huber kernel, the standard robust
defence against outlier loop
closures~\cite{sunderhauf2012switchable}, repairs exactly this
(gated below odometry-only on all $23$) while leaving the ungated
graph far behind ($65.3$ vs.\ $4.9$\,m median): gating and robust
back-ends are complementary in both directions. The ordering
persists across the $\sim\!16\times$ spread of \emph{real}
odometry quality between the three platforms; on $2/23$
St~Lucia-type aliased setups, whose false matches are spatially
close, \method's margin merely narrows.
\emph{The gate's FDR advantage carries through to the
system level (fewer erroneous constraints, lower trajectory
error) with odometry from real sensor streams on all three
platforms, and gating composes with, rather than replaces, robust
back-ends.}

\end{revblock}

\section{Ablation Studies}\label{sec:ablation}

\subsection{Hyperparameter and backbone-size sensitivity}\label{sec:sensitivity}

\Cref{tab:hyperparams} sweeps the two free pipeline parameters
(Mondrian bin count $B$, Lowe ratio $r$) and the \dinov{} backbone
size on the full $23$-setup grid. The headline $23/23$ at
$\alpha=0.10$ is preserved on a plateau ($B \in \{5,8,10\}$, $r \in
\{0.9, 0.95\}$). $B=3$ drops two setups (\fin{coarse bins mix confident and
ambiguous matches, so within-bin shift reaches the accepted part});
higher $B$ values keep $23/23$
at the cost of $\sim\!6$--$10$\,pp TPR (each extra bin \fin{is tested
at confidence $1-\delta/B$ on fewer points}). On the Lowe axis $r=0.95$ ties the default,
$r=0.8$ drops two cells, and the over-strict $r=0.7$ drops to
$19/23$ at $\alpha=0.05$ with $\sim\!16$\,pp lower TPR: the
predictable failure of a ratio so strict that almost no patch matches
survive. Both knobs sit on robust plateaus around our defaults.

\begin{table}[h]
\centering
\caption{Hyperparameter and backbone-size sensitivity on the full
$23$-setup grid. \emph{Top}: Mondrian bin count $B$ (default~$5$);
\emph{middle}: Lowe ratio $r$ (default~$0.9$); \emph{bottom}:
\dinov{} backbone (default ViT-L/14).}
\label{tab:hyperparams}
\scriptsize
\setlength{\tabcolsep}{2.5pt}
\begin{tabular}{l c rrr rrr}
\toprule
& & \multicolumn{3}{c}{$\alpha = 0.05$} & \multicolumn{3}{c}{$\alpha = 0.10$} \\
\cmidrule(lr){3-5}\cmidrule(lr){6-8}
Setting & note & Valid & FDR & TPR & Valid & FDR & TPR \\
\midrule
\multicolumn{8}{l}{\emph{Mondrian bin count} $B$ (Lowe $r = 0.9$, ViT-L)} \\
  $B = 3$           &   & 22/23 & 0.006 & 0.674 & 21/23 & 0.028 & 0.786 \\
  $B = 5$ (default) &   & 22/23 & 0.008 & 0.673 & 23/23 & 0.014 & 0.750 \\
  $B = 8$           &   & 23/23 & 0.002 & 0.618 & 23/23 & 0.008 & 0.684 \\
  $B = 10$          &   & 22/23 & 0.003 & 0.569 & 23/23 & 0.005 & 0.653 \\
\midrule
\multicolumn{8}{l}{\emph{Lowe ratio} $r$ (Mondrian $B = 5$, ViT-L)} \\
  $r = 0.7$           &   & 19/23 & 0.028 & 0.516 & 21/23 & 0.042 & 0.584 \\
  $r = 0.8$           &   & 21/23 & 0.012 & 0.584 & 21/23 & 0.024 & 0.674 \\
  $r = 0.9$ (default) &   & 22/23 & 0.008 & 0.673 & 23/23 & 0.014 & 0.750 \\
  $r = 0.95$          &   & 23/23 & 0.006 & 0.637 & 23/23 & 0.013 & 0.746 \\
\midrule
\multicolumn{8}{l}{\emph{\dinov{} backbone size} (Mondrian $B = 5$, Lowe $r = 0.9$)} \\
  ViT-S/14 (21\,M, $d{=}384$)   & AUROC 0.869 & 21/23 & 0.014 & 0.522 & 22/23 & 0.027 & 0.599 \\
  ViT-B/14 (86\,M, $d{=}768$)   & AUROC 0.916 & 23/23 & 0.009 & 0.594 & 23/23 & 0.017 & 0.687 \\
  \textbf{ViT-L/14 (304\,M)}    & AUROC 0.938 & 22/23 & 0.008 & 0.673 & 23/23 & 0.014 & 0.750 \\
\bottomrule
\end{tabular}
\end{table}

The bottom block sweeps backbone size. ViT-B/14 preserves the
$23/23$ headline at both $\alpha$ levels at $\sim\!2.8\times$ the
throughput, with a $\sim\!6$\,pp TPR drop; indeed it reaches
$23/23$ at $\alpha=0.05$ versus ViT-L's $22/23$, despite $2.2$\,pp
\emph{lower} AUROC (the decoupling again).
ViT-S/14 is another $\sim\!2.4\times$ faster and trades one cell.
ViT-B/14 is the natural embedded starting point.

\emph{Runtime.} The verifier$+$calibrator overhead is $53.1$\,ms
per query on an RTX~5070~Ti (FP16, $T{=}10$ frames, $P{=}256$
patches), dominated by \dinov{} extraction ($2\times26.1$\,ms);
matching and the Mondrian lookup add $<\!1$\,ms, and the overhead is
independent of database size.

\rev{\emph{Selection guideline.} Keep $r=0.9$ and avoid
$r\!\le\!0.8$; choose $B$ as the largest value with
$\gtrsim\!150$ calibration queries per bin (\Cref{sec:limits}).
$B\in\{3,\dots,10\}$ never breaks validity by more than one setup.}

\begin{revblock}
\subsection{Operating limits: target strictness and calibration budget}
\label{sec:limits}

\emph{Stricter targets.} At $\alpha=0.01$ the method degrades by
silence, not FDR violation: on all $23$ setups, for \emph{both}
\method\ and cosine$+$Mondrian, \fin{no threshold passes the}
Clopper--Pearson \fin{screen} and every bin abstains
(TPR $=0$). The cause is resolution, not shift:
$n_\mathrm{cal}\!\approx\!1.4$--$1.9$k over $B=5$ bins cannot
drive the per-bin bound below $\approx\!0.01$. The practical
operating range at these calibration sizes is therefore
$\alpha \gtrsim 0.05$; certifying $0.01$ needs roughly $10\times$
more calibration data per bin.

\emph{Calibration budget.} Subsampling the calibration pool ($10$
seeds per size, all $23$ setups, $\alpha=0.10$): validity is
essentially never lost (worst $98.7\%$ of runs valid); small pools
cost coverage. At $n_\mathrm{cal}\!\geq\!700$
($\approx\!140$/bin) all runs keep non-trivial coverage (TPR
$0.69$); $93\%$ at $400$; $75\%$ at the $5B/\alpha=250$ Mondrian
boundary, below which the vanilla-LTT fallback degrades gracefully
until $n_\mathrm{cal}\!\approx\!50$. \emph{Below $\sim\!150$ calibration queries per bin the system loses coverage, never FDR control; below $\alpha \approx 0.05$ it loses everything to abstention.}
\end{revblock}

\subsection{Negative result: learned re-ranker}\label{sec:learned_reranker}

A cross-attention re-ranker ($118$\,k parameters) gains $+10$ to
$+20$\,pp in-condition R@1 on weak backbones, but its softmax is
valid on only $1/6$ cross-condition setups at $\alpha=0.05$ (mean
FDR $0.137$), and five standard fixes (z-score, temperature scaling,
alternative softmax scores, domain-adversarial training, IWCP) stay
at $1/6$: a score trained with condition information inherits a
condition-dependent distribution.

\section{Discussion and Conclusion}\label{sec:discussion}

\subsection{When does it fail?}\label{sec:failures}

Three failure modes characterise the empirical boundary.

\paragraph{\rev{Repetitive scenery (Nordland): verifier failure with global abstention.}}
Patch matching requires distinguishable local patches. Nordland
(almost entirely forest, snow, and repetitive railway scenery)
leaves \dinov{}-MNN at AUROC $\approx\!0.518$, indistinguishable
from random. The consequence is not false acceptance but total
silence: deployed on Nordland-winter ($3$ backbones, $9657$ queries
each), \fin{no threshold passes the per-bin screen} and the
system accepts $0/9657$ at both $\alpha$ levels.
Four rescue variants (higher resolution, intermediate layers, IDF
re-weighting, cross-frame MNN) either regressed validity
($\sim\!15$\,pp) or added $\sim\!9\%$ FDR. \rev{We report Nordland
as a failure mode: the recipe self-detects uninformativeness, the
FDR target is met only vacuously (TPR $=0$), and the system provides no
localization there.}

\paragraph{\rev{Cross-dataset shift of the low-bin error rate.}}
\rev{Three of $23$ LODO setups fail at $\alpha=0.10$; the quantified
B1-opening mechanism, its directionality, and the target-side
mitigations are analysed in \Cref{sec:robustness}.}

\paragraph{Tight-margin setups.}
The single $\alpha=0.05$ failure (\texttt{nclt\_2013-04-05\_d2vpr})
has been the hardest cell throughout; its empirical FDR sits just
above $0.05$, and a slightly larger calibration pool or
$\alpha=0.06$ would absorb it.

\subsection{Implications, scope, and limitations}

\begin{revblock}
\paragraph{Scope: platforms, viewpoints, and speeds.}
All validation is on ground vehicles with forward-facing cameras and
approximately planar motion. Measured from ground truth, the grid
spans a $\sim\!4\times$ speed range (NCLT's Segway at
$1.1$--$1.2$\,m/s, Oxford at $4.3$--$5.1$\,m/s stop-and-go,
St~Lucia at suburban driving speeds), and empirical validity holds
across it. Sequence matching assumes temporal continuity, so faster
platforms at fixed frame rate see larger inter-frame change;
deployments outside this range should re-tune the sequence length
$T$ rather than assume transfer. Platforms with strong viewpoint
change (aerial, quadruped, humanoid) are outside the validated
scope: large roll/pitch and vertical motion perturb both the
patch-correspondence geometry and the sequence assumption, and per
\Cref{sec:mechanism} the conditional error map cannot be expected to
stay stable under motion regimes never seen in calibration. We would expect coverage loss before FDR violation (the
failure direction observed elsewhere), but this is untested;
on-platform calibration is the safe entry point.
\end{revblock}

\subsection{Conclusion}\label{sec:conclusion}

\method\ is a non-trainable pipeline that recovers empirical FDR
control for sequence VPR under cross-condition deployment by pairing
a frozen \dinov{} patch-MNN verifier with Mondrian conformal LTT.
The central message: raw discrimination does not imply empirical
validity. Equally discriminative and higher-recall scores fall
short on the same calibrator, and a learned re-ranker fares worse. Validity \fin{tracks whether} the bounded patch-MNN ratio keeps a
stable false-match rate inside its accepted region (the quantity
Mondrian's control depends on), not \fin{whether} cal/test distance is
small in any global sense.

The pipeline transfers to $5$\,m metric pose error and to ViT-B/14
for embedded use, \rev{and the gated decision carries to the system
level: in a pose-graph study on real sensor-stream odometry it
suppresses erroneous loop-closure constraints by $29.6\times$ and
cuts median trajectory error by \fin{$9.8\times$}; where
residual false accepts would still hurt, a standard robust kernel
restores gated-below-odometry on all $23$ setups.} \rev{The control is empirical,
not formal: where the score shifts beyond Mondrian's adaptive
capacity (three LODO cells; Nordland), the failure is visible and
diagnosable. Target-side bin-edge adaptation and formal guarantees
under structured shift remain future work.}

\balance
\bibliographystyle{IEEEtran}
\bibliography{references}

\end{document}